\documentclass[3p,times]{elsarticle}

\usepackage{amssymb}

\usepackage[figuresright]{rotating}

\usepackage{subfigure}
\usepackage{xcolor}
\usepackage{amsmath}
\usepackage{hyperref}
\usepackage{array}
\newcolumntype{H}{>{\setbox0=\hbox\bgroup}c<{\egroup}@{}}

\newcommand{\UGS}{UGS}

\makeatletter
\def\ps@pprintTitle{%
 \let\@oddhead\@empty
 \let\@evenhead\@empty
 \def\@oddfoot{}%
 \let\@evenfoot\@oddfoot}
\makeatother

\begin{document}

\begin{frontmatter}




\title{Optimization of the location and design of urban green spaces}


\author[label1]{Caroline Leboeuf}
\author[label1]{Margarida Carvalho}
\author[label2]{Yan Kestens}
\author[label3]{Benoît Thierry}

\address[label1]{CIRRELT and Département d’informatique et de recherche opérationnelle, Université de Montr\'eal}
\address[label2]{Département de médecine sociale et préventive, École de Santé Publique,Université de Montréal}
\address[label3]{Centre de Recherche en Santé Publique, Université de Montr\'eal}

\begin{abstract}
The recent promotion of sustainable urban planning combined with a growing need for public interventions to improve well-being and health have led to an increased collective interest for green spaces in and around cities. In particular, parks have proven a wide range of benefits in urban areas. This also means inequities in park accessibility  may contribute to health inequities. In this work, we showcase the application of classic tools from Operations Research to assist decision-makers to improve parks' accessibility, distribution and design. Given the context of public decision-making, we are particularly concerned with equity and environmental justice, and are focused on an advanced assessment of users' behavior through a spatial interaction model. We present a two-stage fair facility location and design model, which serves as a template model to assist public decision-makers at the city-level for the planning of urban green spaces. The first-stage of the optimization model is about the optimal city-budget allocation to neighborhoods based on a data exposing inequality attributes. The second-stage seeks the optimal location and design of parks for each neighborhood, and the objective consists of maximizing the total expected probability of individuals visiting parks. We show how to reformulate the latter as a mixed-integer linear program. We further introduce a clustering method to reduce the size of the problem and determine a close to optimal solution within reasonable time. The model is tested using the case study of the city of Montreal and comparative results are discussed in detail to justify the performance of the model.
\end{abstract}

\begin{keyword}


Facility Location Problem \sep Mixed-Integer Programming \sep Spacial Interaction Models \sep Urban Green Spaces \sep City Decision-Making \sep Fairness
\end{keyword}

\end{frontmatter}


\section{Introduction}
\label{sec:Introduction}
\paragraph{Context} The need to improve green space coverage and their accessibility in urban areas has been extensively acknowledged in different fields of studies, whether to improve environmental or health conditions. The 2017 WHO report~\citep{who2017urban}, indicated increased attention should be given by policymakers towards urban greening investments. In particular, WHO provides a recommendation on the methods to implement green space interventions in urban settings. These recommendations are built with the intent of improving equity and health, in all of its forms. This is defended by suggesting, for example, to make use of existing collected datasets about green spaces usage and deprivation levels, or to gain a better understanding of the local demographics and parks' users.  

Recent social distancing measures in the context of the Covid-19 pandemic further underlined the perceived value of society towards parks and green spaces. ``The 2021 Canadian city parks report'' \citep{StarkGarrettAberber2021} underlines the Covid-19 context as an additional reason for putting forward discussions relating to parks' accessibility. In that context, various studies emerged with the intent to assess both distribution inequalities and the perceived importance of urban green spaces (\UGS s) in times of crises. Studies and surveys conducted in Belgium \citep{impactCovid19Belgium}, Berlin \citep{UGScovidExperienceBerlin}, China \citep{equitableAccessUrbanParksWuhan}, Madrid \citep{ugsStressCovid19Madrid}, Mexico City \citep{subjectiveWellBeingMexico}, New-York City \citep{UGStimeCrisisNYC} and Poland \citep{impactUGSPoland} unanimously suggest the need for policymakers to invest for improved accessibility to green spaces. 
 Such interventions also contribute to the sustainability of urban developments, and are of particular interest when intense urban densification is observed \citep{HaalandKonijnendijk2015}. 

In this context, we are interested in a model formulation which encompasses common urban planning considerations, while making use of reliable data as an input to recommend a method for improving the distribution and accessibility of parks. As suggested in \cite{who2017urban}, improved accessibility and equity can be attained either through the development of new parks and green spaces {\UGS}s, or improvement of existing ones. 

\paragraph{Problem overview} The question we are interested in the following: how can we leverage existent data and promote equity and environmental justice in city-decisions to improve the accessibility and distribution of {\UGS}s? Based on this premise, we seek to build an optimization model with sufficient flexibility to serve as a template, and can further be modified to account for the attributes that are specific to the green spaces of a city. Such an optimization problem should also be solved in a reasonable time in order to facilitate decision makers' perspectives in terms of experimentation and capacity to compare different scenarios. The model should further account for the decision-making structure in practice while using accurate data. This means resulting decisions with respect to the model's solution should aim to reduce bias, potential for discrimination and subjective motivations. Our work focuses on green spaces as parks. 
Our modeling strategy is based on a two-stage decision process. In the first-stage, the city allocates a budget among the neighborhoods while seeking fairness in terms of existent inequalities. In the second-stage, each neighborhood independently optimizes the usage of urban green spaces by deciding both about their location and design, within the limits of the budget assigned by the city. 

\paragraph{Contributions} Our contributions can be divided in four different axes: \emph{(i)} modeling, \emph{(ii)} a practical case-study, \emph{(iii)} methodological, and \emph{(iv)} experimental.

Our first contribution is the modeling of the two-stage decision-making process in cities for the planning of urban green spaces. To start, we identify the baseline procedure through which cities distribute budgets among neighborhoods. Then, we propose a redistribution of the budgets driven by fairness. Afterward, we model the location of {\UGS}s within each neighborhood. The latter is based on the well-known facility location problem introduced by~\citet{Balinski1965}, with the use of a spatial interaction model to account for individuals' usage patterns of green spaces. More precisely, we modify the standard competitive facility location problem formulation to our context of public facilities provision.

Our two-stage model finds its strength in its flexibility to adapt to different contexts with different model parametrizations. In this line, our contributions are twofold. First, we concretely define the first-stage problem -- fair budget allocation -- using known statistical indexes for inequalities. Second, we generate second-stage instances based on existing datasets. This enables us to test our model on instances reflecting real-world problem topologies.
 
On one hand, the accuracy of our second-stage problem is highly related to the modeling of the demand (i.e., usage of a park by a category of individuals). On the other hand, a granular modeling of the demand results into large second-stage optimization problems. Thus,  our methodological contribution comes in the form of aggregating similar demand points, resulting in smaller problems. Concretely, we propose a clustering technique for the demand points, allowing to reduce the time to solve otherwise large instances, at the cost of a slight reduction in accuracy.  We provide empirical evidence for the  value of our approach to reduce the size of the second-stage problems.
 
Finally, we solve our two-stage model using real datasets available for the city of Montreal. This allows to discuss the effects of budget allocation, the performance of our second-stage size reduction approach, as well as the importance of properly estimating the model parameters for practical use.

To the best of our knowledge, this is the first work that uses a competitive facility location problem in a context of public planning, and that explicitly accounts for inequalities and environmental justice through a two-stage optimization process. Some challenges of this application include the lack of available data for modeling the green space users accurately using statistical techniques. However, we expect our contributions to provide a valuable template for further urban planning models that could assist public decision-makers.

\section{Related work}\label{sec:literature}
 Next, we provide a literature review motivating the studied problem as well as the positioning of our contributions within the existent work. To this end, we start by referring works on {\UGS}s accessibility and inequity, and on the use of spatial optimization to support their planning. Then, we discuss the literature on facility location, particularly focusing on the modeling of user behavior.
 
 \paragraph{Urban green spaces} One of the most popular approaches in assessing UGSs accessibility is to use Geographic Information Systems (GISs). In~\cite{OhJeong2007}, the distribution of parks in the city of Seoul is addressed due to inefficient accessibility. The authors debate that standard statistical indices on the parks' serviceability do not accurately convey whether a park location is in a central or outer area, neither how citizens benefit from them. For instance, many studies use the linear distance to parks, instead of accounting for the population real path choices and travel time, and thus, failing to quantify the population patronizing each park. In addressing this issue, the authors demonstrate the importance of considering factors such as land use and population density around a park. In~\cite{ComberBrundsonGreen2008},~\cite{KabischHaase2014} and~\cite{HoffimannBarrosRibeiro2017},  GISs were also used to assess inequalities of green space accessibility among diverse socioeconomic groups in Leicester, England, in Porto, Portugal, and in Berlin, Germany, respectively. In \cite{CoombesJonesHillsdon2010}, using a GIS database of neighborhood and green spaces characteristics in Bristol, England, the authors show a greater use of {\UGS}s with decreased distance and that {\UGS}s classified as parks tend to increase the physical activity of those living nearby. In \cite{ChiangLiao2011}, the authors develop a GIS-based spatial analysis model for the city of Tainan, Taiwan, to examine in terms of equity the distribution of urban public facilities. In all of these studies, accessibility is deemed necessary when planning the location of UGSs. It should be underlined that in \cite{HoffimannBarrosRibeiro2017}, the authors suggest that conclusions made in a specific location are geographically biased and should not be generalized to other cities. 

Alternative methods were suggested to assess equity in accessibility to UGSs. For example, \citet{NgomGosselinBlais2016} focus in individuals' access to UGSs and suggest refining distance metrics with the use of travel costs and Spatial Interaction Models (SIMs). An ANOVA regression model is developed to explain the distance to green spaces and their total coverage area using significant explanatory variables. Two case studies are used for this purpose, including databases from the cities of Quebec and Montreal, Canada. Regression models show Montreal access to green spaces is less favorable in poorer areas than in wealthier areas. \citet{BooneBuckleyGroveSister2009} recommend a novel approach based on Thiessen polygons to define each park catchment area and on an asymmetric reapportioning of census data, allowing to measure the crowding of a park. They apply the methodology in Baltimore, Maryland, and conclude unfair park access for African Americans. In \cite{Dai2011}, we are introduced to a Gaussian-based \textit{two-step floating catchment area model} to assess spatial accessibility, and to an \textit{ordinary least squares} model to assess socioeconomic disparities in the distribution of UGSs in Atlanta, Georgia. Using geospatial analysis and equity mapping exercises,  \citet{WolchWilsonFehrenbachJed2013} show inequalities between sociodemographic groups in Los Angeles, United States. \citet{Ibes2015} uses a multistep statistical analysis to classify urban parks according to their specific physical and geographic dimensions, as well as their features. This is then mapped to a base layer of social variables in the city of Shanghai, China. In \cite{YeHuLi2018}, the two-step floating catchment area method is used to display changes of UGSs access between 2010 and 2015 in the city of Macau, China.

Spatial optimization is known as a reliable approach in the field of geography and urban planning to help urban planners make decisions about the location of public facilities.  For a detailed review of spatial optimization concepts, we refer  the reader to \cite{Zielinska2016}.  \citet{NeemaOhgai2010} consider the siting of urban parks and open spaces through the case study of the city of Dhaka, Bangladesh. In this paper, the authors formulate a multi-objective model to account for population density, air and noise pollution, and areas with low accessibility to urban parks and open spaces. A genetic algorithm with dynamic weighting is used to solve the optimization problem. In \cite{VallejoCorneVargas2017}, two heuristics are proposed to  optimize the allocation of green spaces over time. In \cite{YoonKimLee2019}, the authors study the optimal location and type of green spaces in a planning model given their greening benefits. The problem defines a multi-objective formulation, maximizing cooling and connectivity, and minimizing costs. The problem is  solved using a non-dominated sorting genetic algorithm.  \citet{YuZhangFuHuangLiCao2020} formulate a multi-objective function to maximize simultaneously the economic, ecological and social value of green spaces and solve the problem using a genetic algorithm. In this paper, the social value acts as the input to model users' demand for UGSs. More recently, \citet{LiLiMa2022} aim to support the decision-making process of UGSs planning using an optimization method that minimizes the land conversion cost of newly added UGS parcels. This solution aims to lower construction costs and improve utility of UGSs. The authors highlight that previous spatial optimization solutions can hardly be used in real-life context, failing to meet the actual constraints of equity and costs. In all these works, the behavior of the population, i.e., their habits in terms of {\UGS}s visits, is not modeled. Moreover, the structure of the decision making process is not included.

\paragraph{Facility location} The problem of location and design of {\UGS}s inherits from concepts of multiple variants of the Facility Location Problem (FLP). In its simpler version, the goal is to determine the location and design of facilities maximizing the number  (or, more broadly, revenue) of individuals who patronize them, subject to a given budget. Thus, in our particular context where facilities are {\UGS}s, modeling the behavior of individuals is extremely relevant for a proper assessment of their accessibility. An extensive stream of studies has made use of the Random Utility Model (RUM)~\cite{McFadden1973} and the proportional (probabilistic) choice rule with deterministic utilities~\cite{Luce1959}. We remark that the term demand is frequently used in this context to mean the proportion of individuals expected to visit a facility.

 Many FLPs studies have made use of the RUM, and authors have provided numerous methods to overcome the non-linearity of the objective function introduced by the use of the RUM. We note the work by \citet{ArosVersMarianovMitchell2013}, in which a mixed-integer linear programming formulation is proposed to model the problem of optimizing the location of park and ride facilities to maximize commuters usage, given a demand that follows a logit model. Motivated by the problem of school network planing, \citet{HaaseMuller2013} propose a method for solving a discrete location model with endogenous demand, in which students' preferences are modeled with a RUM. In \cite{HaaseMuller2014}, a firm's objective is to maximize the frequency of visits, or alternatively, the probability that customers will choose their facility over their competitor's. The objective function is formulated as a mixed-integer non-linear program (MINLP).  In \cite{MullerSvenHaaseKnut2014}, the same authors stretch the importance of customers segmentation to create homogeneous subgroups with shared characteristics at each demand point. \citet{Ljubic2018} and \citet{MaiLodi2020} propose effective exact algorithms for the maximum capture (competitive) FLP when customers demand is modeled through a multinomial logit model. 
 
 The main alternative to RUMs in choice modeling within the framework of competitive FLPs is based on Luce's choice axiom~\cite{Luce1959}, in which the choice probability is defined as a constant ratio of deterministic utilities. Luce's axiom leads to the proportional choice rule, in which customers' probability of choosing a facility is proportional to deterministic utilities \citep{LinTian2021}. The proportional choice rule we are interested in is the SIM, also referred to as the gravity model or Huff-model. As underlined in \cite{BermanKrass1998}, the proportional choice rule is deemed more appropriate for real-life application than the alternative deterministic rule, also referred to as the all-or-nothing approach \citep{AboolianBermanKrass2007b}. For this reason together with their broad applicability (e.g., health-care facility location~\citep{Ammari2000} and electric vehicle charging stations placement~\citep{AnjosGendronMoniz2020}), we focus on SIMs. A stream of studies using competitive FLPs focus on the maximization of market share generated from newly installed facilities. In \cite{AboolianBermanKrass2007b}, the authors introduce the competitive facility location and design problem, where each facility has specific design options in a competitive framework. Here, the goal is to  simultaneously  optimize a facility's location and its design components. The customer's utility is defined as proportional to the facility's attractiveness and to the inverse of the distance. The problem is formulated as an integer non-linear program. In \cite{AboolianBermanKrass2021}, the same authors introduce a more general model, including gravity type and all-or-nothing demand. \citet{DreznerDreznerZerom2018} extend the gravity model by assuming that facilities' attractiveness is random, which is argued as a more realistic approach.

Little attention has been given to inclusion of fairness (or equity) as a component of the FLP. \citet{MarshSchilling1994} made the first detailed literature review of the use of distributive fairness or equity in FLPs, and introduced a unified notation for the measures recognized at the time. \citet{JungKannanLutz2019} underline the limitations of minimizing traditional measures to achieve a fair solution to the classical FLP. They introduce the $\alpha$-fairness measure. Here, the demand is not factored in the problem. \citet{FilippiGuastarobaSperanza2021} study the fair single-source capacitated facility location problem, which presumes that customers will access the provided facilities at an incurred cost. In this paper, the demand is fixed. 

In the 2-stage model presented in the work, we adopt a SIM to reflect the users' behavior and fairness based on available real-world data will be accounted when distributing the city-budget among neighborhoods. 

\section{Model}
\label{sec:Model}

To emphasize on the context of public decision-making, we consider the concept of distributive fairness in the first-stage of the optimization problem. In this step, the budget allocation of a city is distributed among neighborhoods (typically, city administrative subdivisions) such that disadvantaged ones in terms of different socioeconomic and environmental factors are prioritized. The second-stage problem is about the location and (re)design of UGSs within each neighborhood, given its the budget derived in the first-stage process. 

\subsection{First-stage}

The first-stage fair budget allocation step requires distributing a total city budget $B_T>0$ fairly among neighborhoods for UGS planning. Defining concretely fairness depends heavily on the decision-maker's preferred outcome, which varies from one application to another. 
In our practical context, the allocation of budget to neighborhoods by cities can be strongly restricted due to maintenance costs, previous allocation record, etc. Thus, due to the imminent limited freedom to deviate from a pre-determined budget allocation, we will present a simple division of the budget guided by measures of each neighborhood disadvantages.

We denote the set of neighborhoods by $N$. For each neighborhood $n\in N$, the allocated budget is given by the decision variable $b_n$. We further define a fixed baseline budget  $\bar b_n$ for each neighborhood $n \in N$. This value  follows estimation of budget per capita for UGSs investment based on city recommendation, without consideration of other fairness elements. The total city-budget $B_T$ is therefore formulated as
\begin{equation}
    B_T = \sum_n \bar b_n \ .
\label{eq:totalbudget}
\end{equation}
The minimum budget per neighborhood $n \in N$ required to cover the maintenance cost is given by $\underline{b}_n$. The optimization problem consists of maximizing the \emph{fair-weighted city budget} problem~\eqref{Problem:state_1}:
\begin{subequations}
	\begin{alignat}{5}
	&\max_{b \in \mathbb{R}_{\geq 0}^N}  && \sum_{n \in N} b_n\rho_n \label{obj:budget} \\[0.4ex]
		&\mbox{s. t.~~}  &&\sum_{n \in N} b_n \leq B_T     \label{con:totalBudget}\\
		&                && \vert \frac{b_n - \bar b_n}{\bar b_n} \vert \leq \delta \quad \forall n \in N     \label{con:baselineDeviation}\\
		&                &&  b_n \geq \underline{b}_n \hspace{1.4cm} \forall n \in N \ .    \label{con:minimumBudget}
	\end{alignat}
	\label{Problem:state_1}
\end{subequations}
The objective function~\eqref{obj:budget} is guided by the weighting parameters $\rho_n$ for $n \in N$, which act as multiplicative factors, favoring disadvantaged neighborhoods according to predefined attributes. Given the context of UGSs, these fairness attributes include population size, social and material deprivation index, and pollution. Indeed, as reviewed in Section~\ref{sec:literature}, here we can use metrics of accessibility and inequality. Constraint~\eqref{con:totalBudget}  ensures that the total budget of the city is not exceeded. Constraints~\eqref{con:baselineDeviation} enforce the ratio threshold $\delta$ on the maximum neighborhood budget deviation from the baseline budget $\bar b_n$. In this way, we prevent an excessive deviation from the baseline budget.  Constraints~\eqref{con:minimumBudget} guarantee that maintenance costs are at the very least covered.


\subsection{Second-stage}

In the second-stage of the decision process, we take the (independent) perspective of each neighborhood $n \in N$. We underline that here we only consider the territory, parks and population of the neighborhood $n$. The goal is to maximize the overall expected proportion of the neighborhood population visiting its parks. To this end, the neighborhood seeks the optimal (re)design options of existing parks and the optimal location and design of new parks. In the following, for the sake of simplicity, we omit the reference to a neighborhood $n$ and presume that the optimization occurs at the neighborhood level.

We will model our problem as a competitive FLP.  
Here, the neighborhood competes against the no-choice alternative for the users (\cite{BechlerSteinhardtMackertJochen2021}), where they choose not to visit any park due to insufficient attractiveness of the available choices. 

Let $I$ be the set of demand points whose location corresponds to the centroid of the geographic region defining the demand zone.  
The set of segments $S$ allows creating homogeneous categories of individuals with similar usage behavior according to sociodemographic characteristics. In our application, $S$ will define age group segments. Indeed, \citet{SangKnez2016} suggest that children, women and older adults value local parks more importantly given that they tend to spend more time in closer proximity to their home. Hence, for each pair $(i,s) \in I \times S$, we define the weight $w_{is}$ representing the population size percentage located in the demand zone $i$ and in segment $s$ such that $\sum_{i \in I}\sum_{s \in S}w_{is} = 1$. Let $J$ be the set of locations containing the existing and potential new park locations, which are respectively referred to as $\bar{J}$ and $\Tilde{J}$. As for the demand points, facility locations are associated to the centroid of the geographic region defined by an UGS. The distance from demand point $i \in I$ to park $j \in J$, $d_{ij}$, reflects the Euclidean distance between centroid coordinates, and is adjusted to account for approximate walking distance; this will be detailed in the next section. For each location $j \in J$, the set of design options $R(j)$ is specific to it, meaning that the model encompasses a varying number of design options for existing and new parks. Existing parks' baseline design relates to the scenario in which no improvement is made, and maintenance is the only expense to consider. For simplicity, the set of design options equals an ordered sequence of integers, for example $R(j) = [1, \ 2,\ 3]$ signifies that three design options are available at location $j$, where option $1$ is the baseline, option $2$ is an improved and more costly option, and option $3$ is the most expensive option with the most improvement (here, improvement means attractiveness to the users).  Park improvement examples include the addition of park installations, sports fields or children playground, and tree planting \cite{villeDeMontrealConstruction}. For an existing park $j \in \bar{J}$, the associated cost of improving the park with design option $r \in R(j)$  is denoted by $c_{jr}$. Similarly, for new locations $j \in \Tilde{J}$, the notation $c_{jr}$ represents the associated cost of installing a park in the location $j$ with design option $r \in R(j)$. The optimization problem is constrained to a total budget $b$ set by the first-stage allocation.

Next, we describe the utilities of the users (demand points) to derive the SIM and then integrate it in our optimization model. 
 Based in \cite{AboolianBermanKrass2007b}, we define the utility attributed to park $j \in J$ with design $r$ by the demand point $(i,s)$ as follows:
\begin{equation}\label{eq:utilities}
    u_{isjr} = \frac{A_{sjr}}{(1 + d_{ij})^{\beta_{s}}} \ ,
\end{equation}
where the attractiveness parameter is
\begin{equation}\label{eq:attraction}
    A_{sjr} = \alpha_{j} \cdot (1+\theta_{sjr}) \ ,
\end{equation}
and $\alpha_{j} > 0$ is a fix parameter of the baseline attractiveness of $j$ (e.g. walking score), and $\theta_{sjr}$ can be understood as a percentage increase in the attractiveness of demand point's segment $s$ for the design option $r$. The distance decay function $\frac{1}{(1 + d_{ij})^{\beta_s}}$ inherits from the framework introduced by \cite{AboolianBermanKrass2007b}, where $\beta_s$ is the distance sensitivity parameter for age group $s$. We also introduce the parameter $u^0_{is}$, the utility of demand point $i$ in segment $s$ for the no-choice option. In the context of UGSs, we define the following function for quantifying this value:
\begin{equation}
    u^0_{is} = \frac{\frac{1}{|J|}\sum_{j \in J}\alpha_j}{(1+d_{\mathrm{large}})^{\beta_s}} \ ,
\end{equation}
where $\frac{1}{|J|}\sum_{j \in J}\alpha_j$ is the average baseline attractiveness of all park locations, and $d_{\mathrm{large}}$ is a minimum distance threshold value that is deemed too large for someone to want to visit a park (e.g., 1 km). In this way, for parks with low utility, the no-choice option is more attractive. This motivates the optimal UGS planning to install parks and improve parks' design in order to capture more demand that otherwise chooses the no-choice.

We now present the model formulation using the notation and parameters introduced above: 
\begin{subequations}
	\begin{alignat}{5}
	&\max  &&  \sum_{i \in I}\sum_{s \in S} w_{is}\cdot \left(\sum_{j \in J}\sum_{r \in R(j)}p_{isjr} \right)   \label{Obj:ObjectiveLeader_abs15} \\[0.4ex]
		&\mbox{s. t.~~}  &&\sum_{j \in J}\sum_{r \in R(j)}x_{jr} \cdot c_{jr} \leq b   \label{Con:Budget15}\\
		&                &&  \sum_{r \in R(j)} x_{jr} \leq 1 \quad \forall \ j \in J   \label{Con:Design15}\\
		&                &&   \sum_{r \in R(j)} x_{jr} = 1 \quad \forall \ j \in \bar{J}    \label{Con:Installed15}\\
		&                &&    x_{jr} \in \{0,1\} \hspace{0.6cm} \forall \ j \in J, \ \forall \ r \in R(j), \label{Con:binary}
	\end{alignat}
	\label{Problem:stage_2}
\end{subequations}
where the probability of users $(i,s)$ to visit park $j$ with design $r$ is
$$p_{isjr} = \frac{u_{isjr}\cdot x_{jr}}{u^0_{is} + \displaystyle \sum_{k \in J}\sum_{t \in R(k)}u_{iskt}x_{kt}} \ ,$$
and $x_{jr}$ is the location and design decision variable which equals to $1$ if design option $r$ is selected for park location $j$, and 0 otherwise. The objective function~\eqref{Obj:ObjectiveLeader_abs15} corresponds to the total park visits' frequency or, equivalently, to the total market share of the parks controlled by the neighborhood. Constraint~\eqref{Con:Budget15} enforces the budget restriction. Constraints~\eqref{Con:Design15} imply that at most one design option is associated to a park $j$, while Constraints~\eqref{Con:Installed15} guarantee that exactly one design option is selected for an existent park $j \in \bar J$.

In~\ref{app:linearization}, we detail the steps to linearize Problem~\eqref{Problem:stage_2}. After this step, the model can be plugged-in and solved by existent software for mixed integer linear programs.

We conclude this section with a brief discussion of the objective function of the second-stage of the decision-making process. Given the consideration of fairness within the first-stage of the problem, one could also ask about its incorporation in the second-stage. However, there are intrinsic challenges to the consideration of fairness schemes, for instance:
\begin{itemize}
    \item The $L_1$-fairness. For the sake of simplicity, we explain this concept assuming that each demand point patronizes exactly one location or the no-choice. The goal of the $L_1$-fairness objective is to find an UGS plan that minimizes the sum of the absolute difference between the average traveled distance to a park (of all demand points) and the park patronized by each pair $(i,s) \in I \times S$. Alternatively, one could adapt the concept to use utilities instead of distances. We do not consider this objective because we observed in our preliminary tests that the model could provide a \emph{bad} UGS plan: if the average is a large value and the distance to the park patronized by each pair $(i,s)$ is also large, then the $L_1$-fairness can be close to zero.  A possible solution could be to use the $L_2$-fairness where instead of the absolute difference to the average, we consider the $L_2$-norm. However, this would add more non-linearity to the model. Thus, in our experimental results, we will simply analyze the $L_2$-norm metric to evaluate our solutions in terms of the $L_2$-fairness.
    \item The egalitarian fairness. As above, for simplification, we describe this concept  considering that each demand point patronizes exactly one location or the no-choice. Here, the aim is to find an UGS plan that minimizes  the traveled distance to visit the patronized park by the demand point (or pair) traveling the greatest distance to the selected park. Analogously, this concept could be defined in terms of the utilities. In this case, the UGS plan may have to significantly sacrifice all demand points at the expense of optimizing the least favored one. Moreover, since the proportions $w_{is}$, for $(i,s) \in I \times S$, can present a significant variation and that they are not considered in the min-max metric, one may question its fairness. Therefore, we do not present results optimizing this metric, but we use it to evaluate the obtained solutions in the experimental part of this work. 
\end{itemize}

\section{Case study}
The value of our model is shown to the city of Montreal. This choice is motivated  by \emph{(i)} \citet{NgomGosselinBlais2016} who conclude that  Montreal displays more disparity between socioeconomic groups than Quebec city with respect to UGSs accessibility and argue for an improved decision process, and \emph{(ii)} access to data related to green spaces and sociodemographic census. Next, we describe the data available to us and its use to build an instance of our model.

\subsection{Data description}\label{sec:datadescription}
\begin{figure}[!tbp]
  \centering
   \begin{minipage}[b]{0.55\textwidth}\centering
   \includegraphics[width=\textwidth]{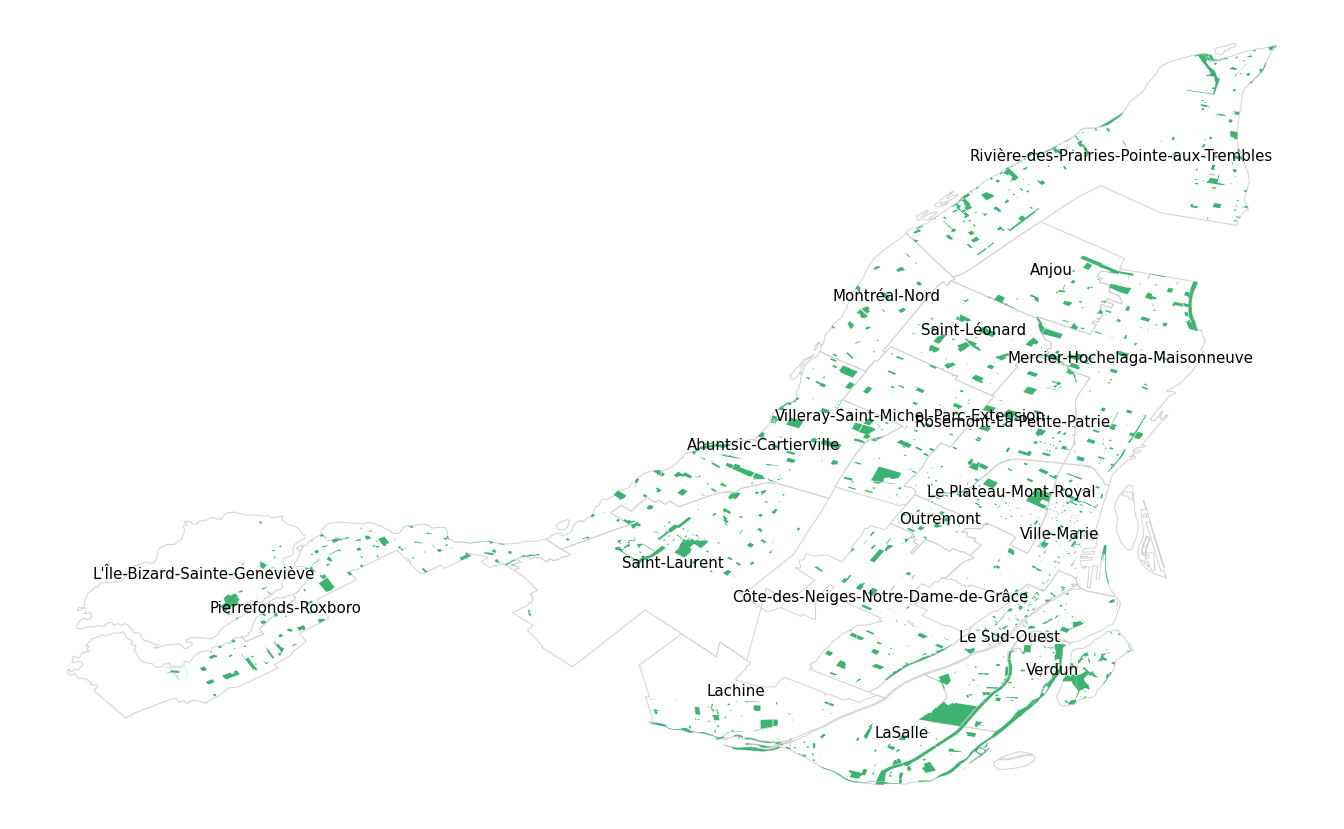}
    \caption{Montreal's boroughs and green spaces}
    \label{fig:montreal}
    \end{minipage}
  \hfill
\begin{minipage}[b]{0.4\textwidth}\centering
    \includegraphics[width=\textwidth]{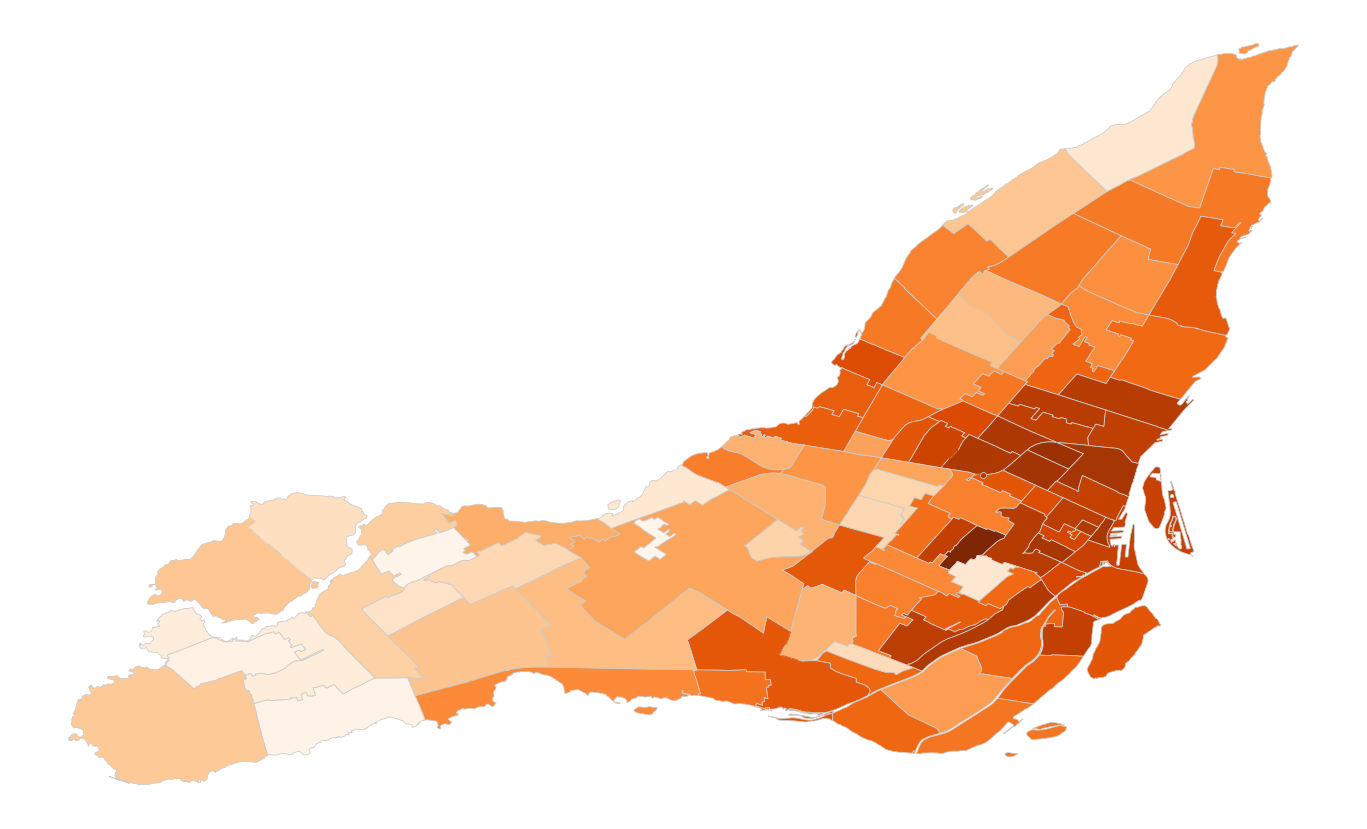}
    \caption{FSA's social deprivation index heatmap. Darker values indicate a higher deprivation}
    \label{fig:social}
  \end{minipage}
\end{figure}

Our set $N$ is composed of the 19 boroughs administrated by the city of Montreal and does not consider the 14 bound municipalities that complete the partition of the island, as they fall under a different jurisdiction. The set of existing parks $\bar{J}$ for each of these boroughs corresponds to the ones managed by it. Therefore, the large parks administrated at the city-level are excluded. We justify this choice by underlining that large parks and neighborhood parks have different objectives, and different usage patterns. Indeed, neighborhood parks are meant to be accessible at a short distance and for daily usage, while large parks can sometimes be accessed with cars and require longer travel time. The dataset used to obtain the existent parks  can be found in~\cite{parcs} and it was complemented by the one in~\cite{installations}. 
Figure~\ref{fig:montreal} displays a map of the city of Montreal with its 19 boroughs, and the borough parks are illustrated with the green zones.

We had also access to the Canadian Active Living Environment (Can-ALE) score.  The Can-ALE is equivalent to the commonly known walking score, and is computed using GIS information collected through Canada. This measure is added with the intention to model the attractiveness of the existing parks.  

Two levels of granularity are considered to define the demand points $I$ for each borough: FSA and postal code. Statistic Canada's 2016 census~ \cite{statcan} provides age group distributions by FSA which results in the definition of $S$ and $w_{is}$ (the population percentage in this demand pair); when $I$ corresponds to postal codes, we assume that $w_{is}$ is equal to the value of its FSA. 

\begin{figure}[!tbp]
  \centering
  \begin{minipage}[b]{0.4\textwidth}\centering
      \includegraphics[width=\textwidth]{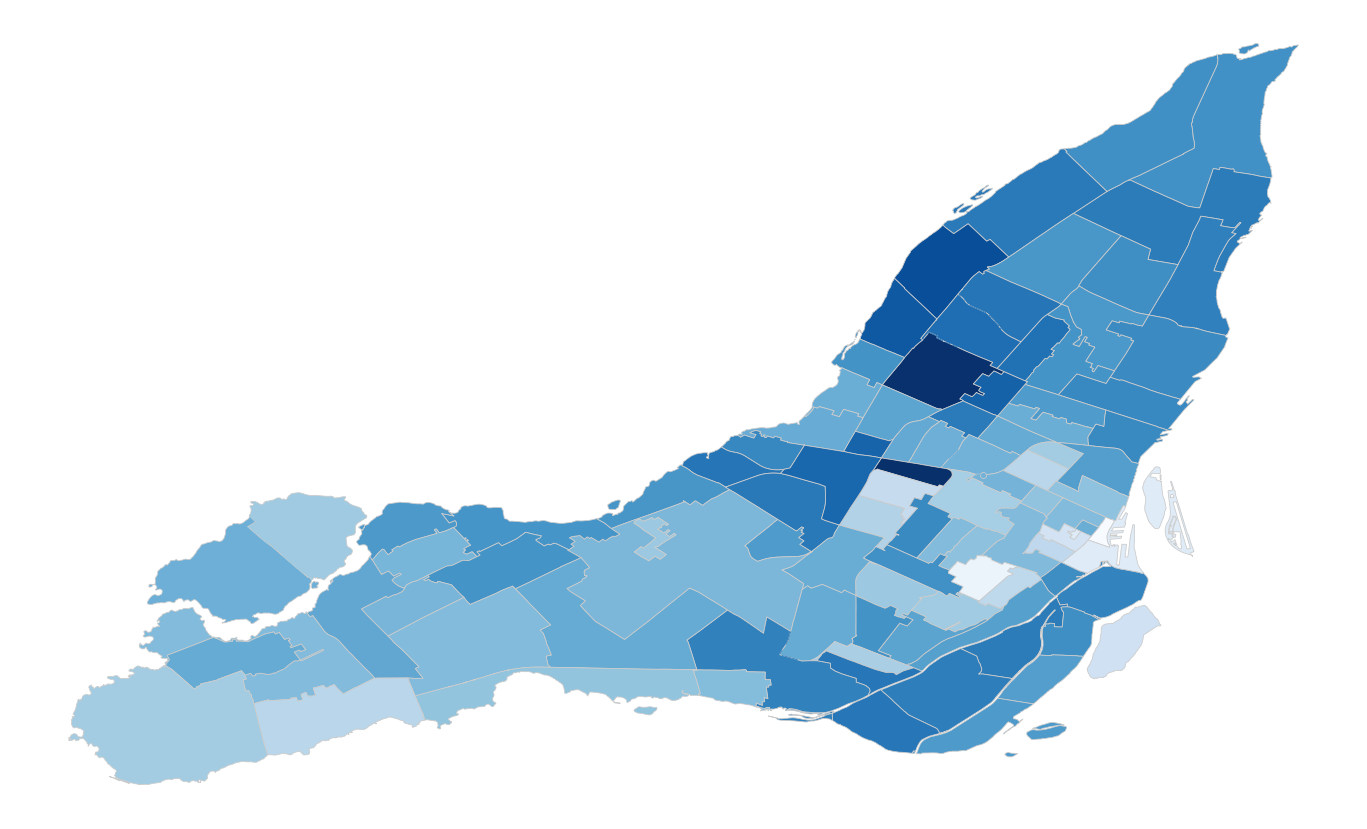}
    \caption{FSA's material deprivation index heatmap. Darker values indicate a higher deprivation}
    \label{fig:material}
  \end{minipage}
  \hfill
  \begin{minipage}[b]{0.4\textwidth}\centering
 \includegraphics[width=\textwidth]{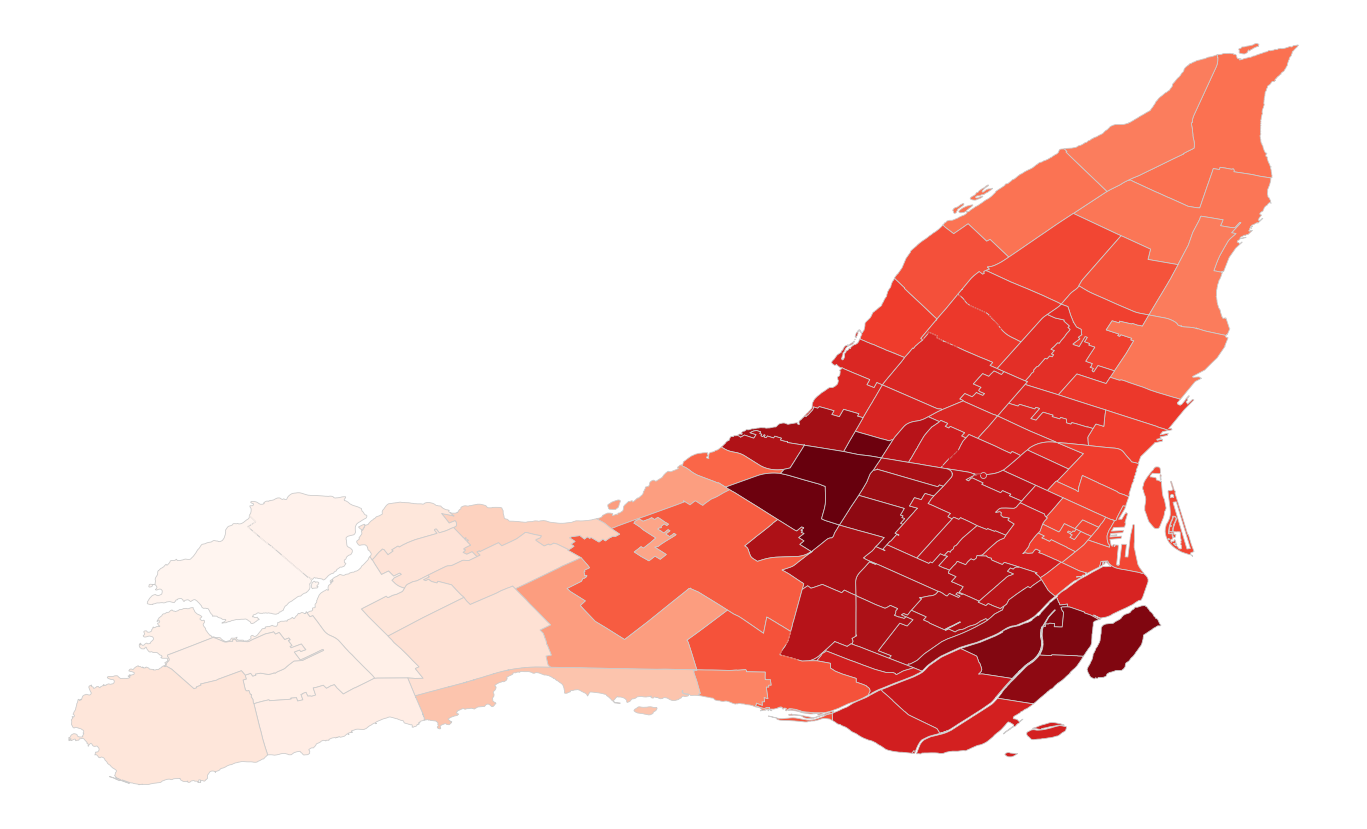}
    \caption{FSA's smoke index heatmap. Darker colors indicate a higher concentration of polluting particles}
    \label{fig:smoke}
  \end{minipage}
\end{figure}

Up to here, we described the essential data elements that will enable us to create the second-stage problems. With respect to the first-stage, we need information to devise the weighting parameters. Thus, we collected the Quebec's Social and Material Indices at the FSA level, also referred to as Pampalon's deprivation indexes~\cite{deprivation}. These measures reflect social and economic inequalities. Figures~\ref{fig:social} and \ref{fig:material} illustrate the value of these metrics. The heatmap shows that the west side of Montreal displays more favorable values with lower deprivation indexes. Finally, a smoke pollution measure was also considered. We used the PM2.5 metric under the Canadian Optimized Statistical Smoke Model, aggregated at the FSA level, but also available at the 6-digit postal code level.  Figure~\ref{fig:smoke} displays the smoke pollution measure, indicating that the denser and more industrial areas of Montreal have higher exposure to air pollution. 

\subsection{Instances}

Our goal is to generate instances mimicking the real-world problem topologies in the city of Montreal. In this way, we can demonstrate the value of our two-stage model formulation to optimize the UGSs location and design, while guaranteeing solvability. We remark that not all parameters of our model can be rigorously determined for our case study due to the lack of data. Namely, the utilities~\eqref{eq:utilities} should be determined through robust statistical techniques. Nevertheless, by taking advantage of the interpretation of the SIM model for the utilities, we discuss the aspects supporting the values considered in our instances.

\subsubsection{First-stage instance}  

The weighting factors $\rho_n$ of neighborhoods $n \in N$ are a result of multiplicative factors for different fairness considerations. The first fairness attribute $\rho_{n1}$ is the population density given its variability as detailed in~\ref{app:demographics}. In particular, in very dense areas, the city should plan on installing more green spaces, given the lack of private gardens. We compare each neighborhood's density to Montreal's average population density, and use this ratio as the multiplicative adjustment, with a maximum adjustment of $\pm 10 \% $ to prevent unbalanced budget allocation. The second attribute is the material deprivation factor $\rho_{n2}$ and the social deprivation factor $\rho_{n3}$. In the same manner as for the population density, we compare each neighborhood's level of material and deprivation index to Montreal's average, and use this ratio as a multiplicative factor with a maximum adjustment of $\pm 5 \% $. We process in the same manner for the fourth fairness component, the smoke pollution factor $\rho_{n4}$. The resulting multiplicative factor $\rho_n$ is set as $\rho_n = \rho_{n1} \times \rho_{n2} \times \rho_{n3} \times \rho_{n4}.$ 
The values of the fairness weighting multiplicative factors are available in~\ref{app:fair}.

The baseline budget is derived from a predefined budget per capita. Therefore, the total yearly baseline budget $\bar{b}_n$ of a neighborhood $n$ varies with respect to the population, and ensures at the very least fairness of budget between individuals. The most recent published budgets of six central neighborhoods are used to estimate the average amount invested in parks per capita for the city of Montreal: Villeray-Saint-Michel-Parc-Extension~\cite{villeray}, Rosemont-La Petite-Patrie~\cite{rosemont}, Montréal-Nord~\cite{nord} and Mercier-Hochelaga-Maisonneuve~\cite{mercier}. We justify this choice given their large size and central location, and also the availability of recent information about the investment made specifically in parks. We use the total park budget, divided by the estimated population of the corresponding neighborhood, as an estimation of the ratio of budget per capita. In this manner, we achieve a final amount of \$42 per capita. Furthermore, since parks' management has a planning horizon over many years, we presume a budget period of five years. Although this model does not yield a solution for each year, it suggests an UGS planning that can be achieved over a period of five years, which is more realistic than a one-year horizon. Hence, for each $n \in N$, $\bar{b}_n$ is determined by multiplying $42\$ \times 5$ by the population size available.\footnote{In~\ref{app:costs}, a potential adjustment of the $\bar{b}_n$ value is discussed so that maintenance costs are guaranteed to be covered.}  Finally, the budget deviation threshold $\delta_n$ is set equal to $0.30$. This value is set with the intention to avoid an unbalanced budget allocation between neighborhoods.

\subsubsection{Second-stage instances}

\paragraph{Demand points} Once the demand points $I$ are set (which can be FSA, postal codes, or clusters as discussed later), each demand point $i \in I$ is segmented in the following age groups $S$: 0-14 years (children), 15-64 years (teenagers and adults) and 65 years or more (elderly). We create these three age classes given an increased sensitivity to distance to parks for younger children and for the elderly. The computation of the corresponding weight $w_{is}$ is determined directly from the data as explained in Section~\ref{sec:datadescription}.

\paragraph{Park (facility) location}The set of existing parks $\bar{J}$ is simply equal to the set of neighborhood parks managed by it. 
The set of new parks $\Tilde{J}$ is simplified by assuming a potential new location at each FSA centroid. In real-life application, the decision-maker has knowledge about locations that can be converted in UGSs, and their exact location. Our approximation allows us to cover a neighborhood almost uniformly, with a reasonable number of locations.  Concerning design options, for each location $j \in J$, we consider three fix alternatives ($|R(j)|=3$). 

\paragraph{Costs}  Regarding the approximation of the costs of park management we use the report on Montreal's 2020 performance indicators~\cite{indicator}. This allow us to set a maintenance cost per $m^2$ and account for a proportional cost increase depending on the design option $j$:
$$c_{jr} = c_{\mathrm{maintenance}} \times area_j \times (1+0.8 \cdot (r-1)).$$
We refer the reader to~\ref{app:costs} for a detailed description on the determination of the costs $c_{jr}$.

\paragraph{SIM model} The basic attraction level $\alpha_j$ in Equation~\eqref{eq:attraction} is set equal to the ALE score of location $j \in \bar{J}$ for already installed parks, and to the ALE score of FSA $j$ for a new park $j \in \Tilde{J}$. The attraction increase $\theta_{sjr}$ for existing parks $j \in \bar{J}$ is set equal to a percentage increase in line with the suggested design options:
$$\theta_{sjr}=0 \textrm{ for } r=1, \quad \theta_{sjr}=0.5 \textrm{ for } r=2, \quad  \theta_{sjr}=1 \textrm{ for } r=3.$$
The attraction increase for design options $r \in R(j)$ is fixed for all age segments $s \in S$ and facility points $j \in J$, given the lack of data on park usage by age. Nevertheless our model
 allows for more realistic parameters. We note that $\theta_{sj1}$ is set equal to 0 because it is associated to the maintenance option $r = 1$. The attraction increase for a new park $j$ is set equal to:
$$\theta_{sjr}=0.75 \textrm{ for } r=1, \quad \theta_{sjr}=1.5 \textrm{ for } r=2, \quad  \theta_{sjr}=3 \textrm{ for } r=3.$$
As noted, we presume a large attraction increase for newer parks, given individuals' preference for new installations. This assumption can be refined to account for long term decrease of attraction.\

The distance sensitivity parameter $\beta_s$ is set according to different age groups. This parameter should be estimated using robust statistical methods introduced in \cite{Huff2003}, but are roughly approximated in our research, given the limited availability of park usage data. For the children and elderly age groups, we set $\beta_s$ to 1.5, and to 1.0 for teenagers and adults. Indeed, as suggested earlier, children have a higher sensitivity to distance given that park visits are usually supervised by parents, or, regarding children's independent mobility, are often restricted close to home, while elderly require closer facilities given reduced mobility. \

Finally, since our optimization model presumes non-negative utilities, to prevent errors in the solving step, we shift all utilities $u_{isjr}$ to positives values. Indeed, negative utility values can exist due to negative Can-ALE scores. Thus, to maintain the scale of utilities, we simply shift them instead of normalizing them. 

\paragraph{Distance} The distance $d_{ij}$ from demand point $i \in I$ to existing parks $j \in \bar{J}$ is calculated using the Euclidean distance formula, adjusted with a multiplicative factor of 1.3 to approximate the actual travel distance. This adjustment factor is derived using an average of the Google Distance Matrix distances divided by the corresponding Euclidean distances for a set of points. For new parks, the distance from the demand point belonging to the same FSA as the park is assumed equal to 500 meters, and 1000 meters for demand points outside the FSA where the park is located. In real-life application context, one needs to compute the actual travel distance for the selected demand point to the exact new park location. Given that the location of new park is set at the centroid of the FSAs for simplicity, we opt for this assumption. Next, given the context of the UGSs planning, we define a maximum distance threshold. Indeed, according to the literature, neighborhood parks should be located close enough to residential areas so that individuals have a minimum incentive to visit them~\cite{pedestrians}. For this reason, we set the maximum distance to parks for children at 500 meters. For all other age groups, maximum distance for parks less than 50,000 m$^2$ is set to 500 meters, and to 800 meters for larger parks. Indeed, people have more incentive to walk longer for larger UGSs. In the model, we set the utility $u_{isjr}$ to 0 for distances above these predefined thresholds.

\section{Experiments}

 In this section, we do not  prescribe an UGSs planning for the city of Montreal, but rather demonstrate the flexibility of our two-stage optimization model and prove its applicability and performance using a large real-life based instance.
 
 In Section~\ref{subsec:clustersmethod}, we present our clustering methodology to solve the second-stage problems efficiently without losing much accuracy.  In Section~\ref{subsec:results}, we demonstrate the value of our clustering methodology for the second-stage problems as well as the analysis of the results from solving the first-stage problem. The details on our computational setup can be found in~\ref{app:setup}.   
 
 In what follows, we present tables of results. The meaning of the columns in these tables are as follows:  ``GAP (\%)'' provides the optimality gap as a percentage, ``RunTime (s)'' provides the time in seconds to solve an optimization problem, ``ObjVal (\%)'' provides the value of the best objective computed by the solver. Finally, the discussed objective values relating to Objective~\eqref{Obj:ObjectiveLeader_abs15} and discussed in the following sections should be interpreted as the percentage of the population that is using parks. Therefore, the aim is to be close to 100\%.\

\subsection{Clusters method}\label{subsec:clustersmethod}
 The city of Montreal has 97 FSA and 44,117 unique 6-digit postal codes. Even though Problem~\eqref{Problem:stage_2} is solved for each borough, the consideration of postal codes as the demand points $I$ will lead to prohibitively large problems.  
 Thus, we also consider the clustering of postal codes to obtain the set of demand points and reduce computational time. We perform this clustering through the well-known \textit{k-means} method using their latitude and longitude coordinates. 
 
 We validate our methodology with a concrete example. After careful analysis of the sociodemographic and geographic attributes of the neighborhoods and their corresponding parks' distribution and accessibility, we consider the neighborhood Rosemont-La Petite-Patrie to test the different aggregation levels of the set of demand points. This borough is one of the largest of the central boroughs, having the second-largest population density in the city of Montreal, and displaying a relatively low ratio of park area per capita. Using a neighborhood of such large size, with a potential for large number of demand points, allows us to establish if the second-stage model can perform well with large instances. In particular, it serves our purpose of identifying the right balance between demand granularity and model solvability.

The three aggregation levels considered are the postal codes, clusters of postal codes, and FSAs. Our baseline neighborhood has a total of 2,331 postal codes, but only a total of five FSAs. Thus, the computation time resulting from the postal codes' method is expected to be unreasonably large. Therefore, we suggest the clusters' method to address the over-simplified FSA basis and the over-complex method using the postal codes. Using the cluster methodology described above, we obtain 200 clusters in the neighborhood of Rosemont. Each cluster is presumed to be located at its centroid, namely the mean latitude and longitude of the postal codes belonging to the cluster. The number of clusters is chosen such that a significant reduction in clustering runtime is reached for a specific number of clusters.  
 
 Table~\ref{tab:rosemont} displays the performance of the solver for the three problem versions. It also provides the objective value for the FSA and cluster solutions for the problem using $I$ as postal code as well as the objective value for the FSA solution using  $I$ as the clusters. Clearly, the FSA problem version is much easier to solve (4 seconds), while the other versions with more granular demand points reach the time limit, meaning that we obtain a potentially suboptimal solution. However, the evaluation of the FSA solution in the two problems with more granular demand points suggests a potential overestimation done by the FSA aggregation level objective, and that the postal code method increases the percentage of population's park visits by 0.5\% compared to the FSA solution. Using the clustering method, the problem is still slow to solve. Despite that, within the time limit of one hour, and provides a better objective (82.9\%) to the postal code problem. Therefore, we manage to increase the postal code's result from 75.7\% to 82.9\% using a methodology that requires one hour instead of five hours. 
 

\begin{table}
\centering
    \begin{tabular}{ l H r r r  }
    \hline
    Method & Status & GAP (\%) & RunTime (s) & ObjVal (\%) \\
    \hline
    FSA & 2 & 0.0 & 4.00 & 84.0 \\
    Postal Code & 9 & 14.8 & 18,000.00 & 75.7 \\
    Clusters & 9 & 2.4 & 3,600.00 & 83.4 \\ \hline
    Postal Code Problem with FSA solution &  &  &  & 75.2 \\
    Postal Code Problem with cluster solution &  &  &  & 82.9 \\
    Clusters Problem with FSA solution &  &  &  & 75.1 \\
    \hline
    \end{tabular}
    \caption{Model results for Rosemont-La Petite-Patrie with baseline budget}
\label{tab:rosemont}
\end{table}

 For each neighborhood, the number of predefined clusters is given in~\ref{app:clusters}.
 
 \subsection{Results}\label{subsec:results}
 
Next, using the results in Tables~\ref{tab:clusters_baseline} and~\ref{tab:clusters_fair}, we compare the fair budget allocation with the baseline budget allocation. Namely, \emph{(i)} we analyze how the available budget per neighborhood affects the optimal value of the second-stage objective,  \emph{(ii)} we discuss the performance of our approach, \emph{(iii)} we demonstrate that the adoption of a SIM goes beyond the use of distance as an element to guide UGS planning through the $L_1$-norm metric. In \ref{app:nochoice}, we discuss the importance of properly estimating the no-choice utility. 
 

\begin{table}[!ht]
\centering\footnotesize
    \begin{tabular}{p{7cm} r H r r r r}
    \hline
    Neighborhood & Budget (M\$) & Status & GAP (\%) & RunTime (s) & ObjVal (\%) &  L$_1$-norm\\
    \hline
    Ahuntsic-Cartierville 	&	33.3	&	9	&	0.7	&	3600	&	99.0	&	1,164,820	\\
    Anjou 	&	9.5	&	2	&	0	&	29	&	93.8	&	189,600	\\
    Côte-des-Neiges-Notre-Dame-de-Grâce 	&	27.3	&	9	&	2.5	&	3600	&	82.1	&	31,602	\\
    L'Île-Bizard-Sainte-Geneviève 	&	6.4	&	2	&	0	&	12	&	95.9	&	249,174	\\
    Lachine 	&	10.4	&	2	&	0	&	1880	&	97.5	&	243,233	\\
    LaSalle 	&	17.2	&	9	&	1.2	&	3600	&	97.3	&	269,140	\\
    Le Plateau-Mont-Royal 	&	18.6	&	9	&	1.2	&	3600	&	94.4	&	34,906	\\
    Le Sud-Ouest 	&	17.7	&	9	&	0.1	&	3600	&	99.0	&	368,463	\\
    Mercier-Hochelaga-Maisonneuve 	&	28.9	&	9	&	0.7	&	3600	&	99.0	&	447,265	\\
    Montréal-Nord 	&	17.8	&	2	&	0	&	1313	&	66.7	&	28,229	\\
    Outremont 	&	6.6	&	2	&	0	&	28	&	84.3	&	32,268	\\
    Pierrefonds-Roxboro 	&	19.1	&	9	&	0.5	&	3600	&	98.5	&	504,732	\\
    Rivière-des-Prairies-Pointe-aux-Trembles 	&	33.1	&	9	&	0	&	3600	&	99.4	&	634,183	\\
    Rosemont-La Petite-Patrie 	&	29	&	9	&	2.4	&	3600	&	83.4	&	34,258	\\
    Saint-Laurent 	&	22.4	&	9	&	0.3	&	3600	&	98.4	&	328,378	\\
    Saint-Léonard 	&	15.7	&	2	&	0	&	3068	&	63.0	&	23,032	\\
    Verdun 	&	36.2	&	2	&	0	&	2457	&	70.9	&	21,577	\\
    Ville-Marie 	&	18.8	&	9	&	3.7	&	3600	&	92.7	&	30,888	\\
    Villeray-Saint-Michel-Parc-Extension 	&	30.4	&	9	&	3	&	3600	&	83.1	&	31,135	\\
    \hline
    Average & & & & & 89.2 \\
    \hline
    \end{tabular}
    \caption{Model results using the clusters method and baseline budget}
\label{tab:clusters_baseline}
\end{table}

\begin{table}[!ht]
\centering \footnotesize
    \begin{tabular}{p{7cm} r H r r r r}
    \hline
    Neighborhood & Budget (M\$) & Status & GAP (\%) & RunTime (s) & ObjVal (\%) &  L$_1$-norm\\
    \hline
Ahuntsic-Cartierville 	&	33.3	&	9	&	0.7	&	3600	&	99	&	 1,164,820 \\
    Anjou 	&	8.3	&	2	&	0	&	90	&	93.5	&	 196,100 \\
    Côte-des-Neiges-Notre-Dame-de-Grâcece 	&	35.5	&	9	&	1.7	&	3600	&	84.2	&	 33,325 \\
    L'Île-Bizard-Sainte-Geneviève 	&	6.4	&	2	&	0	&	12	&	95.9	&	 249,174 \\
    Lachine 	&	10.4	&	2	&	0	&	1877	&	97.5	&	 243,177 \\
    LaSalle 	&	12.1	&	9	&	0.8	&	3600	&	97.2	&	 274,292 \\
    Le Plateau-Mont-Royal 	&	24.2	&	9	&	0.8	&	3600	&	94.9	&	 35,506 \\
    Le Sud-Ouest 	&	16.2	&	9	&	0.6	&	3600	&	99	&	 369,811 \\
    Mercier-Hochelaga-Maisonneuve 	&	26.3	&	9	&	0.7	&	3600	&	99	&	 447,360 \\
    Montréal-Nord 	&	12.4	&	2	&	0	&	1285	&	63.8	&	 27,052 \\
    Outremont 	&	4.6	&	2	&	0	&	32	&	80.7	&	 28,502 \\
    Pierrefonds-Roxboro 	&	19.1	&	9	&	0.5	&	3600	&	98.5	&	 504,732 \\
    Rivière-des-Prairies-Pointe-aux-Trembles 	&	33.1	&	9	&	0	&	3600	&	99.4	&	 634,183 \\
    Rosemont-La Petite-Patrie 	&	33.3	&	9	&	4.1	&	3600	&	83.2	&	 34,084 \\
    Saint-Laurent 	&	21.4	&	9	&	0.7	&	3600	&	98.4	&	 330,393 \\
    Saint-Léonard 	&	13	&	2	&	0	&	625	&	56.9	&	 19,023 \\
    Verdun 	&	36.2	&	2	&	0	&	2164	&	70.9	&	 21,577 \\
    Ville-Marie 	&	13.1	&	9	&	1.9	&	3600	&	91.6	&	 30,124 \\
    Villeray-Saint-Michel-Parc-Extension 	&	39.5	&	9	&	1.6	&	3600	&	84.3	&	 31,656 \\

    \hline
    Average &  &  &  &  & 89.0 &\\
    \hline
    \end{tabular}
    \caption{Model results using the clusters method and fair budget}
\label{tab:clusters_fair}
\end{table}

\begin{figure}[t]
    \centering
    \includegraphics[scale=0.2]{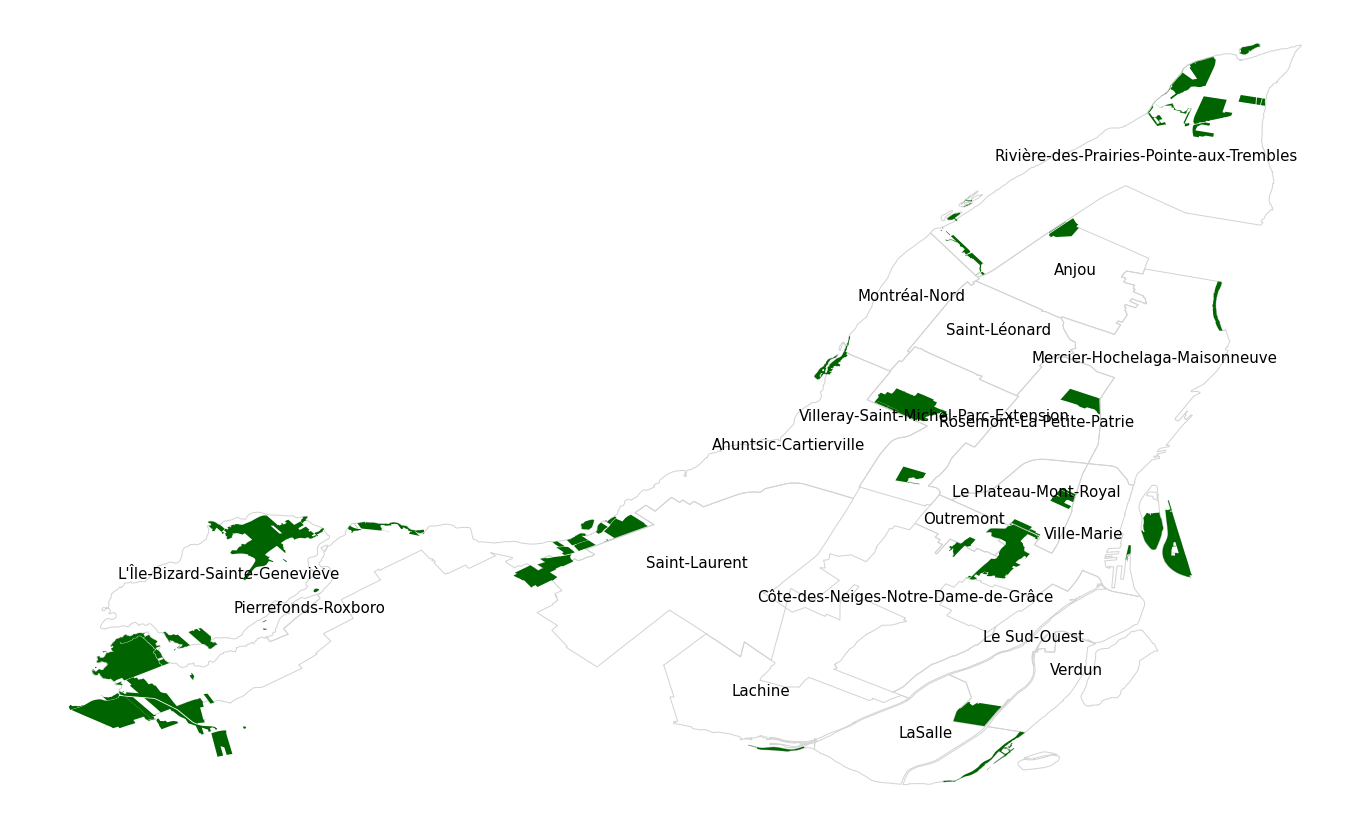}
    \caption{Large parks managed by the city of Montreal}
    \label{fig:large_parks}
\end{figure}

\paragraph{Objective value}
The neighborhoods that benefit the most from a budget increase with the fair budget allocation method include Côte-des-Neiges-Notre-Dame-de-Grâce and Villeray-Saint-Michel-Parc-Extension, with an objective increase of 2.1\% and 1.2\% respectively. These boroughs are deemed very large residential areas, and differ from other neighborhoods mostly due to higher population density. We underline one of the model's shortcomings here, in that we ignore the large parks of the city of Montreal in our problem. Indeed, although these neighborhoods do display a lack of borough's parks when compared to wealthier neighborhoods such as Outremont or more remote and greener neighborhoods such as Pierrefonds-Roxboro or L'Île-Bizard-Sainte-Geneviève, they do have access to large parks including Frédéric-Back Park, Jarry Park and Mont-Royal Park to name a few. In Figure~\ref{fig:large_parks}, we display the large parks that are managed at the city-level and that are excluded of the optimization problem.

As a result, the suggested model is deemed sufficiently flexible to include these considerations should it be done. Again, the purpose of this work is not to make an UGS planning recommendation, but to prove the applicability and the performance of our methodology. The neighborhoods displaying the largest sacrifice under the fair budget allocation are Montréal-Nord, Outremont and Saint-Léonard, with a respective objective value decrease of 2.8\%, 3.7\% and 6.1\%. These neighborhoods have the largest budget decrease, and suffer from a budget cut at the cost of multiple neighborhoods which are already at the minimum tolerable budget to cover the maintenance cost.

We further question the model as to how does the fair allocation method impact the probability of individuals visiting parks. Tables~\ref{tab:clusters_baseline} and \ref{tab:clusters_fair} display the average objective value weighted with the population of each neighborhood. The average totals to 89.2\% for the baseline budget, and to 89.0\% for the fair budget, which suggests a slight decrease of the overall objective under the fair allocation method. This should be expected given that the budget is re-allocated to neighborhoods with higher deprivation, to act as a compensation with improved green spaces coverage. Therefore, this reduces UGS budgets of neighborhoods that are greener, to be able to invest more in deprived ones.


\paragraph{Performance}
Regarding performance, around 40\% of the second-stage optimizations are solved to optimality within the time limit, and the highest optimality gap is 4.1\%. The most demanding neighborhoods with respect to total runtime are Ahuntsic-Cartierville, Mercier-Hochelaga-Maisonneuve and Ville-Marie, which have the largest number of postal codes, and thus, of clusters. As expected, we observe no monotonic relation between the budget and the optimal GAP (a proxy of the problem difficulty). This is expected since the second-stage model size (i.e., number of variables and constraints) does not depend on the budget. 

\paragraph{$L_1$-Norm}  
We now analyze how the fair allocation affects the average distance from the demand points to the park locations from the final solution. To this end, we use the \textit{$L_1$-Norm} on the expected distance from individuals' to the surrounding parks using the probabilities $p_{isjr}$; the formal definition is provided in~\ref{app:l1}. Lower values of the \textit{$L_1$-Norm} would suggest an increase of fairness, with lower deviation from the average distance to parks. These results actually show that for some neighborhoods with an increase of budget when using the fair allocation, such as Côte-des-Neiges-Notre-Dame-de-Grâce, Le Plateau-Mont-Royal and Villeray-Saint-Michel-Parc-Extension, we observe an increase of the \textit{$L_1$-Norm}. This counterintuitive observation shows that the use of utilities instead of distances to model the preferences of the population significantly changes the nature of the problem. As an example, Outremont is found with a significant decrease of budget with the fair allocation, but also results in a lower \textit{$L_1$-Norm} measure. This stretches the importance of the definition of the neighborhoods' objective in the decision process and the (unsurprising) sensitivity of the solution according to the modeling of park visits. \\

\paragraph{Final remarks}
For a sample solution, \ref{app:rsmt} displays the design options selected under the baseline and the budget allocation method in the neighborhood of Rosemont. As noted in these results, new parks hypothetically located at the centroid of the FSAs are also subject to change of design options when subject to a change of neighborhood budget. Based on the previous results, we can safely suggest that the model yields reasonably realistic results, even though the parameters' are based on assumptions from existent statistical indexes. Refining such assumptions, and thus, parameters estimation, should improve the credibility of urban planning recommendations based on the model's results.

\section{Conclusions and perspectives}

To our knowledge, this is the first work simultaneously \emph{(i)} considering a city decision-making process in sequence, intrinsic to administrative subdivision structure, \emph{(ii)} including the concept of fairness through the city's budget-allocation to neighborhoods (subdivisions), and \emph{(iii)} modeling the population access to facilities administrated by each neighborhood  using SIMs. This work serves as a proof-of-concept to the use of our model as a decision making tool for the planning of UGSs by cities.

One of the main shortcomings in our case-study comes from the lack of use of data to use statistical techniques for estimating the parameters of the choice model. Refining such parameters would improve the solution's accuracy. Another shortcoming of the model comes from the consideration of long-term planning. Indeed, we considered a static model using a total budget estimated for a period of five years, but future work should consider an extension of this model to yield a budget planning per year, and potentially account for a dynamic response at each period, i.e., a change on demographics and also on the utilities of the SIM.  Future work should also focus on improving the process for defining the exact number of clusters required to achieve optimal results under an acceptable time limit and gain knowledge on the trade-offs between the exact method and the clusters' method should significantly improve the reliability of the results. We also suggest a comparison of the different decision-making structures. Indeed, in the city of Montreal, decisions are taken in a two-stage fashion. It would be interesting to compare this process to the alternative where the city also takes the second-stage decisions; in this way, we could see the advantages and limitations of a coordinated second-stage. This approach would substantially increase the size of the model, which could be addressed with the use of the clustering technique. 

Our work also opens new perspectives, namely, the integration of green paths in the model. These paths are intended to connect parks to each other and display a sufficient amount of trees and vegetation to be considered as a ``green'' path. Another addition is to account for blue spaces, defined as the space allocated to water bodies or watercourses. In the 2021 WHO report~\citep{braubach2021green}, green spaces are considered simultaneously with blue spaces as a mean to address more recent concerns about climate change and mental health \citep{braubach2021green}.

We believe this work can contribute to the stream of literature that puts forward methodologies for transforming theoretical concepts into practical decision making tools,  to the benefit of environmental initiatives, public health and equity.

\section*{Acknowledgements}

The authors wish to thank Kia Babashahi and Nurit Oliker for preliminary discussions on decision-making model.  

This work was funded by FRQ-IVADO Research Chair in Data Science for Combinatorial Game Theory, and the Planning and Dissemination Grant - ICS (FRN: 164998) from CIHR.

\appendix
\section{Linearization of problem~\eqref{Problem:stage_2}}
\label{app:linearization}

Using the ``Method‑Based Linearization'' technique reviewed in \cite{BechlerSteinhardtMackertJochen2021}, we introduce the following non-negative auxiliary variables
$$v_{is} = \frac{1}{u^0_{is} + \sum_{j \in J }\sum_{r \in R(j) } u_{isjr}x_{jr}} \quad \forall i \in I, \forall s \in S \ ,$$
with $v_{is}$ taking values in the interval $$\left[\frac{1}{u^0_{is} + \sum_{j \in J }\sum_{r \in R(j) } u_{isjr}}, \ \frac{1}{u^0_{is}}\right] \ .$$

The objective function of problem~\eqref{Problem:stage_2} becomes
\begin{equation}
     \sum_{i \in I}\sum_{s \in S} w_{is}\left(\sum_{j \in J }\sum_{r \in R(j) } u_{isjr}x_jv_{is}\right) \ ,
    \label{Obj:ObjectiveLeader_abs15_2}
\end{equation}
and we add the constraint
\begin{equation}
    p^0_{is} + \sum_{j \in J}\sum_{r \in R(j) }x_{jr}p_{isjr} = 1
\end{equation}

$$ \iff u^0_{is}v_{is} + \sum_{j \in J}\sum_{r \in R(j) }u_{isjr}x_{jr}v_{is} = 1 \ ,$$

where $$p^0_{is} = \frac{u^0_{is}}{u^0_{is} + \sum_{k \in J}\sum_{t \in R(k)}u_{iskt}x_{kt}} \ .$$

In this formulation, we still remain with variable multiplication (bilinear terms) $x_{jr}\cdot v_{is}$. Proceeding with the following re-expression of variable multiplications,
$$z_{isjr} = x_{jr}\cdot v_{is} \quad \forall i \in I, \forall s \in S, \forall j \in J, \forall r \in R(j) \ ,$$
we obtain the resulting mixed integer linear program:
\begin{subequations}
	\begin{alignat}{5}
	&\max  &&  \sum_{i \in I}\sum_{s \in S} w_{is}\left(\sum_{j \in J }\sum_{r \in R(j) } u_{isjr}z_{isjr}\right)
    \label{Obj:ObjectiveLeader_abs15_3} \\[0.4ex]
		&\mbox{s. t.~~}  && \eqref{Con:Budget15}-\eqref{Con:binary} \nonumber\\
		&                &&  u^0_{is}v_{is} + \sum_{j \in J}u_{isjr}z_{isjr} = 1 \hspace{0.6cm} \forall j \in J\\
		&                &&  z_{isjr} \geq 0 \hspace{3.6cm} \forall i \in I, \forall s \in S, \forall j \in J, \forall r \in R(j) \\
		&                &&  z_{isjr} \leq v_{is} \hspace{3.4cm} \forall i \in I, \forall s \in S, \forall j \in J, \forall r \in R(j) \\
		&                &&  z_{isjr} \leq K_{isjr(1)}x_{jr} \hspace{2.4cm} \forall i \in I, \forall s \in S, \forall j \in J, \forall r \in R(j)\\
		&                &&  z_{isjr} \geq v_{is} + K_{isjr(2)}(x_{jr}-1) \ \ \  \qquad \forall i \in I, \forall s \in S, \forall j \in J, \forall r \in R(j)\ , 
	\end{alignat}
	\label{Problem:stage_2_MILP}%
\end{subequations}
where $K_{isjr(1)}$ and $K_{isjr(2)}$ are sufficiently large numbers. In this case, we will set it to the maximum upper bound of $z_{isjr} =x_{jr} \cdot v_{is} \leq v_{is} \leq \frac{1}{u^0_{is}}=K_{isjr(1)}=K_{isjr(2)}$.

\section{Neighborhood statistics}\label{app:demographics}
Table~\ref{tab:analysis} presents statistics for the 19 neighborhoods of the city of Montreal. It shows variability between urban neighborhoods, underlining the need for fair public decision-making. One aspect to bear in mind here is that only neighborhood parks are considered. For this reason, some special considerations should be made in areas where very large natural parks exist, such as L'Île-Bizard-Sainte-Geneviève, Pierrefonds-Roxboro and Rivière-des-Prairies-Pointe-aux-Trembles. Such considerations are made a posteriori in our analyzes of the empirical results.
 \begin{table}
\centering \footnotesize
    \begin{tabular}{lrrrrrr}
    \hline
     & Area &  & Density  & Average  &Percentage  & Ratio of park  area\\
    Neighborhood & (km$^2$) & Population & (km$^2$) &  Income (\$) & of park area &   per capita (m$^2$) \\
    \hline
       Ahuntsic-Cartierville   	&	24.3	&	135000	&	5600	&	70000	&	8.5	&	 15.3 \\
    Anjou   	&	13.9	&	42810	&	3000	&	69000	&	3.7	&	 11.9 \\
    Côte-des-Neiges-Notre-Dame-de-Grâce   	&	21.5	&	166000	&	7700	&	68000	&	3.7	&	 4.8 \\
    L'Île-Bizard-Sainte-Geneviève 	&	23.7	&	18000	&	800	&	115000	&	1.7	&	 22.5 \\
    Lachine 	&	17.9	&	44600	&	2500	&	70000	&	4.6	&	 18.4 \\
    LaSalle 	&	16.4	&	77000	&	4700	&	64000	&	6.5	&	 14.0 \\
    Le Plateau-Mont-Royal 	&	8.1	&	104000	&	12800	&	67000	&	8.6	&	 6.7 \\
    Le Sud-Ouest 	&	15.8	&	79000	&	5000	&	66000	&	12.4	&	 24.7 \\
    Mercier-Hochelaga-Maisonneuve 	&	25.5	&	136000	&	5300	&	60000	&	7.3	&	 13.7 \\
    Montréal-Nord 	&	11	&	83000	&	7600	&	52000	&	3.3	&	 4.4  \\
    Outremont 	&	3.8	&	24000	&	6300	&	175000	&	4.5	&	 7.2 \\
    Pierrefonds-Roxboro 	&	27.2	&	69000	&	2500	&	87000	&	5	&	 19.5 \\
    Rivière-des-Prairies-Pointe-aux-Trembles 	&	42.5	&	108000	&	2500	&	75000	&	4.8	&	 19.1 \\
    Rosemont-La Petite-Patrie 	&	15.9	&	140000	&	8800	&	64000	&	6.1	&	 6.9 \\
    Saint-Laurent 	&	43.1	&	98000	&	2200	&	81000	&	3.1	&	 13.4 \\
    Saint-Léonard 	&	13.6	&	79000	&	5800	&	64000	&	6.2	&	 10.6 \\
    Verdun 	&	9.7	&	69000	&	7100	&	85000	&	23	&	 32.3 \\
    Ville-Marie 	&	16	&	88000	&	5500	&	73000	&	3.7	&	 6.7 \\
    Villeray-Saint-Michel-Parc-Extension 	&	16.5	&	143000	&	8700	&	54000	&	5.4	&	 6.2 \\
    \hline
    \end{tabular}
    \caption{Neighborhood data analysis}
\label{tab:analysis}
\end{table}

\section{Weighting fairness parameters} \label{app:fair}

The factors below are used to derive the final weighting fairness parameter $\rho_n$ for each neighborhood $n \in N$:

$$\rho_n = \rho_{n1} \times \rho_{n2} \times \rho_{n3} \times \rho_{n4} \ .$$

\begin{table}[!ht]
\centering \footnotesize
\begin{tabular}{r r r r r r }
    \hline
    Neighborhood $n$ & Density $\rho_{n1}$ & Social $\rho_{n2}$ & Material $\rho_{n3}$ &  Smoke $\rho_{n4}$ & $\rho_n$ \\
    \hline
    Ahuntsic-Cartierville   & 0.90 & 0.95 & 1.05 & 1.00 & 0.90 \\
    Anjou   & 1.03 & 0.95 & 1.05 & 1.00 & 1.02 \\
    Côte-des-Neiges-Notre-Dame-de-Grâce   & 1.10 & 1.05 & 0.95 & 1.01 & 1.11 \\
    L'Île-Bizard-Sainte-Geneviève & 0.90 & 0.95 & 0.95 & 0.96 & 0.78 \\
    Lachine & 0.90 & 1.05 & 1.05 & 0.99 & 0.98 \\
    LaSalle & 0.90 & 0.95 & 1.05 & 1.01 & 0.90 \\
    Le Plateau-Mont-Royal & 1.10 & 1.05 & 0.95 & 1.00 & 1.10 \\
    Le Sud-Ouest & 0.90 & 1.05 & 0.95 & 1.01 & 0.90 \\
    Mercier-Hochelaga-Maisonneuve & 0.90 & 1.05 & 1.05 & 0.99 & 0.99 \\
    Montréal-Nord & 1.10 & 0.95 & 1.05 & 1.00 & 1.09 \\
    Outremont & 1.10 & 0.95 & 0.95 & 1.01 & 1.00 \\
    Pierrefonds-Roxboro & 0.90 & 0.95 & 1.05 & 0.97 & 0.87 \\
    Rivière-des-Prairies-Pointe-aux-Trembles & 0.90 & 0.95 & 1.05 & 0.99 & 0.89 \\
    Rosemont-La Petite-Patrie & 1.10 & 1.05 & 0.95 & 1.00 & 1.10 \\
    Saint-Léonard & 1.10 & 0.95 & 1.05 & 1.00 & 1.10 \\
    Saint-Laurent & 0.92 & 0.95 & 1.05 & 1.01 & 0.92 \\
    Verdun & 0.90 & 1.05 & 0.95 & 1.02 & 0.91 \\
    Ville-Marie & 1.10 & 1.05 & 0.95 & 1.00 & 1.10 \\
    Villeray-Saint-Michel-Parc-Extension & 1.10 & 1.03 & 1.05 & 1.01 & 1.19 \\
    \hline
    \end{tabular}
\end{table}

\section{Park costs}\label{app:costs}

In~\cite{indicator}, the total amount related to managing parks is established at \$80 per capita, or \$31,000 per hectare. For simplicity, we use the estimated cost per hectare and convert it to the equivalent cost of \$3.10 per m$^2$ of park land. For new parks, we approximate the acquisition and installation cost to a total of \$15 per m$^2$. This gross estimation deserves refinement, given that acquisition prices are very sensitive to location and size. For this reason, our model presumes a hypothetical price, but it can be replaced with an appropriate amount depending on the locations of potential new UGSs. As a side note, the neighborhood baseline total budget is adjusted in exceptional cases where the budget per capita is insufficient to cover the minimal maintenance costs. This scenario happens when the neighborhood has a large park coverage. To address this, we set the baseline neighborhood budget as the maximum of the budget per capita formula, and the minimal budget required for maintenance adjusted with a 1.05 factor. The decision of adding a 5\% budget increase to the minimal maintenance budget is to ensure that areas in the neighborhood with lower green spaces accessibility have the opportunity for new installations.  In real-life applications, urban planning involves a  varying number of design options according to each location, with an estimated budget according to each scenario. Here, we suggest a more simplistic approach. The associated cost $c_{jr}$ for each design option is defined as follows:
$$c_{jr} = c_{\mathrm{maintenance}} \times area_j \times (1+0.8 \cdot (r-1)),$$
where $c_{\mathrm{maintenance}}$ is the maintenance cost of \$3.10 per m$^2$, $area_j$ is the size of the park in m$^2$, and $(1+0.8 \cdot (r-1))$ is a multiplicative factor that increases the total cost of each design option with an additive percentage of 80\%. For new parks located at the FSA centroid, we presume a land area of 50,000 m$^2$, which is equivalent to a large neighborhood park. As a reminder, the first design option for an existing park is the basic maintenance scenario where no improvement is made, while for new parks, the first scenario corresponds to the smallest investment option. If no new park is to be located at the location $j \in \Tilde{J}$, then the variable $x_{jr}$ would be equal to 0. For existing parks, constraints of the model formulation ensure that at least one design option is selected. 

\section{Experimental Setup} \label{app:setup}

In this section, we review the experimental setup used in the results presented in this paper. To start, we note that our methodologies are implemented in Python 3.9.7. Optimization models are solved with Gurobi 9.5.0 using 6 cores. The experiments run on a Dual-Socket Intel(R) Xeon(R) Gold 6226 clocked at 2.70GHz (12 Cores per Socket, 24 Cores total) and is equipped with 376 GB RAM. Below, we describe the predefined model's parameters, and how to interpret the different model's attributes. The details below are specific to the second-stage optimization, given that the first-stage consists of a fairly simple linear program, and it does not require additional model parametrization. \

First, we introduce the different model's parameters as defined in Gurobi \footnote{\url{https://www.gurobi.com/documentation/9.5/refman/parameter_descriptions.html}}: the pre-solve level, the degenerate simplex moves limit, and the time limit. For the clusters and the postal codes' methodologies, namely the larger instances, the pre-solve level is set to the option 2, the maximum value available, which implies a much longer and complex pre-solving step but whose preliminary results indicated to compensate due to tighter final  optimality gaps. The degenerate simplex moves limitation parameter is set to 0, as recommended in Gurobi for problems resulting in an important number of \textit{Total elapsed time} messages in the log. These choices were made to speed up the computing time and follow Gurobi's guidelines in this context. We also set the time limit for solving the UGS planning in each neighborhood to 3,600 seconds (one hour) for both the FSA and the clusters' methodology, and to 18,000 seconds (five hours) for the postal codes' method. The reason for these time limits is to ensure that the application of our solution can easily be used in a context of city-decision making and that multiple tests can be done in reasonable time.  These different time limits according to the demand aggregation level is simply to adapt the computing time depending on the problem size, although it is known that the time to solve an NP-hard problem is expected to increase exponentially with its instance size.

Next, Gurobi's key model's attributes\footnote{\url{https://www.gurobi.com/documentation/9.5/refman/attributes.html}} are discussed to compare methodologies: the runtime, the GAP, and the objective value. While the first-stage model is a simple linear program, the second-stage program solved for each neighborhood is a mixed-integer linear program, which can be slow to solve. Regarding runtime, for smaller neighborhoods, Gurobi's solver is able to reach an optimal solution with a small runtime, but larger instances usually reach the predefined time limit. In the latter case, we simply retrieve the best solution found by the solver within the time limit. The GAP is defined as the absolute value of the gap percentage between the objective value reached by the best feasible objective value found so far and the tightest bound computed, and serves as a good indicator of the improvement of the solution at each iteration over the branch-and-bound process. Therefore, when the solver reaches optimality, the final GAP is 0, while in the opposite case, it is strictly greater than 0; the latter will occur for runs where the time limit is reached. 


\section{Clusters details} \label{app:clusters}
The number of clusters for each neighborhood is presented in Table~\ref{tab:clusters}.
\begin{table}[!ht]
\centering \footnotesize
    \begin{tabular}{l r}
    \hline
    Neighborhood & Number of clusters \\
    \hline
    Ahuntsic-Cartierville   & 250 \\
    Anjou   & 100 \\
    Côte-des-Neiges-Notre-Dame-de-Grâce   & 150 \\
    L'Île-Bizard-Sainte-Geneviève & 50 \\
    Lachine & 100 \\
    LaSalle & 150 \\
    Le Plateau-Mont-Royal & 150 \\
    Le Sud-Ouest & 200 \\
    Mercier-Hochelaga-Maisonneuve & 250 \\
    Montréal-Nord & 150 \\
    Outremont & 50 \\
    Pierrefonds-Roxboro & 150 \\
    Rivière-des-Prairies-Pointe-aux-Trembles & 200 \\
    Rosemont-La Petite-Patrie & 200 \\
    Saint-Laurent & 200 \\
    Saint-Léonard & 100 \\
    Verdun & 100 \\
    Ville-Marie & 250 \\
    Villeray-Saint-Michel-Parc-Extension & 200 \\
    \hline
    \end{tabular}
\caption{Number of clusters for each neighborhood}
\label{tab:clusters}
\end{table}

\section{\texorpdfstring{$L_1$} \ -norm of the distance per neighborhood} \label{app:l1}

The $L_1$-norm for a set of demand points $I$ and segments $S$ is defined as follows:
$$L_1 = \sum_{i \in I}\sum_{s \in S}w_{is}\cdot |\bar{d}_{is} - \bar{d}|, $$
where
$$\bar{d}_{is} = \sum_{j \in J}\sum_{r \in R(j)} p_{isjr}d_{ij} $$
and
$$\bar{d} = \frac{\sum_{i \in I}\sum_{s \in S} w_{is}\bar{d}_{is}}{\sum_{i \in I}\sum_{s \in S}w_{is}}.$$

\section{No-choice option}\label{app:nochoice}
In this section , we discuss the impact of the quantification of the no-choice utility parameter $u^0_{is}$ on the scale of the objective value. Indeed, the model's results show multiple neighborhoods with significantly high probability values near 100\%, including in Ahunstic-Cartierville, Le Sud-Ouest, Mercier-Hochelaga-Maisonneuve and Rivière-des-Prairies-Pointe-aux-Trembles. Although these residential areas can evidently display a higher propensity for park visits, this could also be a consequence of the underestimation of the no-choice utility and this, reflects the importance of parametrizing a model with proper data estimation. To support this idea, Table~\ref{tab:u0} exemplifies how the objective value varies with respect to different no-choice utility in the neighborhood of Outremont (with the baseline budget). We chose this neighborhood (instance) since we can compute optimal second-stage solutions to it within the time limit and, thus, properly evaluate the sensitivity to the no-choice utility. The sensitivity analysis is made by applying a multiplicative factor to the no-choice utility parameter, $u^0_{is}, \ \forall \ i,s \ \in \ I,S$ as indicated in Table~\ref{tab:u0}.\\

\begin{table}[!ht]
\centering
    \begin{tabular}{r r}
    \hline
    Multiplicative factor & ObjVal (\%)\\
    \hline
    110\% & 83.4 \\
    100\% & 84.3 \\
    90\% & 85.3 \\
    80\% & 86.4 \\
    70\% & 87.5 \\
    \hline
    \end{tabular}
    \caption{No-choice utility sensitivity in the neighborhood of Outremont}
\label{tab:u0}
\end{table}

\section{Design option solutions} \label{app:rsmt}

Table~\ref{table:existent} and Table~\ref{table:newparks} provide the optimal design solutions for  existent and new parks, respectively.

\begin{table}[!ht]
\centering
\begin{tabular}{r r r}
    \hline
    Park ID & Baseline budget & Fair budget \\
    \hline
    0082-000 & 1 & 1 \\
    0082-001 & 2 & 2 \\
    0126-000 & 1 & 1 \\
    0127-000 & 3 & 1 \\
    0187-000 & 1 & 1 \\
    0190-000 & 3 & 3 \\
    0196-000 & 2 & 1 \\
    0197-000 & 1 & 1 \\
    0200-000 & 3 & 2 \\
    0202-000 & 1 & 1 \\
    0204-000 & 3 & 3 \\
    0205-000 & 3 & 3 \\
    0206-000 & 1 & 1 \\
    0207-000 & 1 & 1 \\
    0209-000 & 1 & 1 \\
    0210-000 & 1 & 1 \\
    0211-000 & 1 & 1 \\
    0212-000 & 1 & 1 \\
    0216-000 & 3 & 3 \\
    0217-000 & 3 & 3 \\
    0218-000 & 3 & 2 \\
    0219-000 & 1 & 1 \\
    0220-000 & 1 & 1 \\
    0221-000 & 2 & 1 \\
    0241-000 & 1 & 1 \\
    0243-000 & 3 & 3 \\
    0244-000 & 3 & 1 \\
    0291-000 & 3 & 2 \\
    \hline
    \end{tabular}
    \quad
    \begin{tabular}{r r r}
    \hline
    Park ID & Baseline budget & Fair budget \\
    \hline
    0292-000 & 3 & 1 \\
    0300-000 & 1 & 3 \\
    0303-000 & 1 & 1 \\
    0589-000 & 3 & 3 \\
    0590-000 & 1 & 1 \\
    0785-000 & 1 & 1 \\
    0878-000 & 1 & 1 \\
    0879-000 & 2 & 1 \\
    1017-000 & 3 & 3 \\
    1037-000 & 1 & 1 \\
    1038-000 & 1 & 1 \\
    1064-000 & 1 & 1 \\
    1133-000 & 1 & 1 \\
    1166-000 & 1 & 1 \\
    1167-000 & 1 & 1 \\
    1168-000 & 1 & 1 \\
    1169-000 & 1 & 1 \\
    1170-000 & 1 & 1 \\
    1171-000 & 2 & 1 \\
    1172-000 & 2 & 2 \\
    1173-000 & 1 & 1 \\
    1174-000 & 3 & 1 \\
    1186-000 & 3 & 1 \\
    1209-000 & 3 & 3 \\
    1272-000 & 3 & 3 \\
    1273-000 & 3 & 1 \\
    7000-000 & 1 & 1 \\
     &  &  \\
    \hline
    \end{tabular}
    \caption{Optimal design solutions for existent parks}
    \label{table:existent}
\end{table}

\clearpage

\begin{table}[!ht]
\centering
\begin{tabular}{r r r}
    \hline
    Park FSA & Baseline budget & Fair budget \\
    \hline
    H1T & 3 & 3 \\
    H1X & 3 & 0 \\
    H1Y & 1 & 3 \\
    H2G & 1 & 3 \\
    H2S & 1 & 1 \\
    \hline
    \end{tabular}
    \caption{Optimal design solutions for new parks}
    \label{table:newparks}
\end{table}




\bibliographystyle{elsarticle-num-names}
\bibliography{bibliography}







\end{document}